# An Evolutionary Computing Enriched RS Attack Resilient Medical Image Steganography Model for Telemedicine Applications

Romany F. Mansour[1] and Elsaid MD. Abdelrahim[2]

[1]Department of Mathematics, Faculty of Science, New Valley, Assiut University Egypt
[2]Department of Mathematics Computer Science Division, Faculty of Science, Tanta University, Tanta, Egypt
romanyf@aun.edu.eg, elsaid_abdelrahim@yahoo.com

***Abstract*—The recent advancement in computing technologies and resulting vision based applications has given rise to a novel practice called telemedicine that requires patient diagnosis images or allied information to recommend or even perform diagnosis practices being located remotely. However, to ensure accurate and optimal telemedicine there is the requirement of seamless or flawless biomedical information about patient. On the contrary, medical data transmitted over insecure channel often remains prone to manipulated or corrupted by attackers. The existing cryptosystems alone are not sufficient to deal with these issues and hence in this paper a highly robust reversible image steganography model has been developed for secret information hiding. Unlike traditional wavelet transform techniques, we incorporated Discrete Ripplet Transformation (DRT) technique for message embedding in the medical cover images. In addition to, ensure seamless communication over insecure channel, a dual cryptosystem model containing proposed steganography scheme and RSA cryptosystem has been developed. One of the key novelties of the proposed research work is the use of adaptive genetic algorithm (AGA) for optimal pixel adjustment process (OPAP) that enriches data hiding capacity as well as imperceptibility features. The performance assessment reveals that the proposed steganography model outperforms other wavelet transformation based approaches in terms of high PSNR, embedding capacity, imperceptibility etc.**

## I. Introduction

The high pace emergence in technologies and associated computing approaches have given rise to a broad horizon serving an array of applications among which medical image communication is a dominating one. In present day health care system, biomedical image transmission has become one of the inevitable needs to ensure efficient seamless and secure data transmission over uncertain channels. Presently, the telemedicine field has gained impressive momentum which has enabled efficient and timely diagnosis remotely. However, the probable adversaries, caused due to (image) information deviation, could not be ignored. In most of the cases, the key biomedical image information is transmitted distinctly in a perceptible manner. No doubt, such transmission often remains prone to the attack within channel and hence there can be compromise towards the secrecy of the patient's key details. To maintain seamless medical image communication, Digital Imaging and Communications in Medicine (DICOM) has developed significant standard for medical data storage, information extraction, printing and transmission over channels. The efficacy of DICOM has enabled it a potential model for medical data communication. No doubt, DICOM files can be easily communicated between two parties which are authorized and capable of receiving image and associated patient details in DICOM format. However, DICOM doesn't have any specific security feature, as it merely stores the data in unqualifiable (or non-editable) manner. On the contrary, there are many application scenarios where images originated from medical examinations might require sharing inadvertently across stakeholders. Thus exposing significant information. This key information could be accessed by any party without having access right and this as a result could cause misuse. Interestingly, in telemedicine applications ensuring intact image attribute and key information is inevitable. In such applications retaining image information intact during transmission throughout channel is a must. With this motivation, a number of efforts have been made to enable data security during transmission. However, with specific medical image information security data hiding or data embedding approaches have gained widespread attention. Various data hiding approaches such as watermarking, cryptography or cryptosystems and steganography have experienced exponential application to secure the data. Among these dominant technologies, steganography is more efficient and suitable for critical data security over uncertain channel conditions [1]. Steganography deals with hiding critical information onto the cover image, which is then followed by transmission over unsecured channel [1]. The secure information or secret message could be text, audio, video, or image data. The robustness of steganography enables it to be applied for various medical imaging purposes, watermarking, inno-cipher, Geo-Static Information (GSI) etc; where medical imaging system deals with hiding significant patient related information in allied medical images like X-Rays, Ultrasounds, CT scans, and MRIs.

Ensuring optimal steganography demands imperceptibility [2] that comprises the following:

**Invisibility-**Since the efficacy of steganography lies in the ability to be unnoticed, invisibility is stated to be the first and foremost inevitable need of the steganography technique.

**Capacity-**Steganography intends to ensure optimal data security while ensuring higher embedding (secret message volume) capacity. Typically, it is measured as bits per pixel (bpp) unit.

**Resilience against Statistical attacks and manipulation-**It signifies the volume of alteration the stego medium or the cover image can endure or resist before certain adversary could obliterate the embedded or the hidden critical information.

To perform data hiding in cover images or medical images, numerous efforts have been made [3]. These approaches are broadly classified into two types of techniques, spatial domain and frequency domain techniques [4], where the first approach embeds the secret data directly into the least significant bit (LSB) of the cover image; while in the latter approach the secret information is embedded in the LSB of transform coefficients. Unfortunately, in the time domain approaches the quality of the stego images often gets distorted and thus, making it vulnerable to get attacked due to perceptibility issues. On the other hand, frequency domain approaches are confined due to low resilience towards attacks [5]. Perceptibility caused by embedding, is the prime reason behind such limitations, and hence to deal with it, authors [6] suggested optimal pixel adjustment process (OPAP) based LSB embedding technique. This technique [6] exhibited better stego-image quality with relatively lower computational complexity. Observing the fundamental process of medical data security, especially by means of steganography, ensuring minimal perceptibility and higher embedding capacity is must, and to achieve it enhancing image transformation for message embedding is a must. Unlike major existing approaches, where wavelet transforms have been applied, in this paper, the ripplet transform (RT) technique has been applied that plays vital role in retrieving or selecting most significant coefficient for data hiding. In addition, it can alleviate the existing issues of the general discrete wavelet transform (DWT) called two dimensional (2D) singularity problem [7]. Being a reversible steganography, we apply inverse discrete ripplet transform (DRT) for data extraction at the receiver end. One more significant novelty of the proposed work is the use of enhanced evolutionary computing (EC) scheme called Adaptive Genetic Algorithm (AGA) for block mapping to be applied for OPAP purpose. This as result not only ensures imperceptibility but also ensures higher embedding capacity. One of the key novelties of the presented research work is the implementation of dual security model (DSM) where in addition to the enhanced DRT based steganography; RSA cryptosystem has been applied to secure the data access. The efficacy of the proposed approach has been investigated in conjunction with statistical attack assessment over stego image, where it has exhibited higher PSNR, low entropy, minimal or even negligible perceptibility and histogram variations.

The other sections of the presented manuscript discuss the following. Section II discusses the related work, which is followed by the discussion of the proposed discrete Ripplet transform (DRT) Type-III and its application for enhanced steganography technique for medical data security. This section also discusses the implementation model proposed and associated technologies such as DRT based image transformation and data hiding, AGA based block mapping and OPAP, RSA cryptosystem based dual security model. Section IV discusses the results and analysis and the overall conclusion and future scopes are presented in Section V. References used in the research work are presented at the end of the manuscript.

## II. Related Work

This section presents some of the key works discussing medical data security using steganography (say, medical image steganography (MIS) and various associated technologies.

In recent years, the significance of telemedicine has been realized globally and hence it has emerged as one of the dominating research domains across academia-industries. However, to ensure seamless and flawless processes, maintaining optimal medical data security is a must. Steganography, being one of the most efficient approaches for medical data security applied image transformation schemes to embed critical data embedding within cover image (i.e., medical image). In fact, the efficiency of reversible steganography techniques significantly relies on the efficacy of image transformation, data embedding and pixel adjustment to enable maximum imperceptibility. In most of the existing approaches integer wavelet transform techniques (IWT) or wavelet transformation approaches have been applied [8][9][10][11] for steganography. In [8], authors developed areal-time data embedding model using IWT technique where it was applied in the transform domain. Being an image compression based approach authors have applied IWT as it outputs in the integer form and hence consumes low memory space. In [12], authors applied IWT based steganography for medical image security. Authors focussed on converting multiple medical images into single one where the cover image was processed wit left-flipping and a dummy cover image was generated. Authors considered the patient's medical diagnosis image as the secret image and to retrieve the scrambled image they applied Arnold transform. In process, the scrambled medical diagnostic image (i.e., secret image) was embedded into the dummy cover image and IWT was applied to obtain the dummy secret image. In the next phase, authors fused the cover image with the dummy secret image to obtain stego-image. However, the computational overheads of such techniques can't be ignored.

Authors in [10], derived a reversible data hiding or critical data embedding approach using compressive-steganography technique. Authors applied wavelets transform technique to exhibit data embedding; however could not address the distortion caused due to compression and its effect on the hiding capacity. To further enhance compression based reversible steganography for medical data security over uncertain transmission channel, a low distortion reversible data hiding scheme was developed in [13], where authors compressed a fixed section of the signal, which is prone to get distorted. Tian [11] applied pixel value difference expansion approach to enable a high capacity reversible data embedding for image steganography. However, authors could not address the aftermath consequences leading vulnerability towards attacks. Similarly, in [14] [15] a Difference Expansion (DE) embedding model was applied where authors exploited HAAR wavelet transform by using horizontal as well as vertical difference images to perform secret data hiding. Authors [15] found that DE along with sophisticated location map with enhanced expandability could achieve higher embedding capacity. With goal to exploit histogram based approaches for steganography, authors [16] applied a histogram shifting approach that shifts (by one pixel) a fraction of the histogram parts in between the peak and the zero level to the right direction. This as a result created an empty bin in conjunction with the peak point where they hide data. Similarly, in [17], a histogram modification model was incorporated where authors modified histogram on pixel differences that similar to [16] forms sufficient place to hide data. However, these approaches could not address the attack conditions, especially statistical attacks in channel.

Authors [18] focused on medical data confidentiality issue through steganography where the cover images was at first transformed into one-dimensional sequence by means of Hilbert filling curve, which was then processed for splitting into non-overlapping clusters of three pixels in each. To enrich imperceptibility, authors applied adaptive pixel pair match (APPM) based data embedding, where pixel value differences (PVD) of the three pixels individually is retrieved and data is embedded in those pixel ternaries. This as a result causes low distortion and hence high imperceptibility.

Considering efficacy of LSB embedding, authors [19] developed a LSB matching model where data hiding was performed in the edges and the pixel bits of the secret message were matched with the LSB plane. In [20] a fuzzy logic-based steganography model was proposed for medical diagnostic image security. Authors applied random LSB selection based approach to hide the secret data. Authors applied personal data and the diagnostic suggestions as the secret information that was compressed and encrypted to enable attack resiliency. However, this approach might be complicated. In [21] authors derived a data hiding model for 3D MRI images, where at first they applied segmentation to perform region localization followed by LSB embedding. Authors [22] applied two-phase contourlet transform technique for data hiding, where at first the cover image was split into non-overlapped blocks and the secret information was embedded in the high frequency component. In addition, they embedded the secret data in the low frequency component of the global contourlet transform. Authors [23] derived a digital steganography model to hide Electronic Patient Records (EPR) into medical diagnosis images without making any significant modification in the image part. Realizing the fact that the human visual system (HVS) is less sensitive to changes in high contrast image sections, Authors [23] exploited edge detection technique to recognize and embed secret data in spiky image-parts. They applied Hamming code to embed three distinct secret message bits into 4 bits of the cover image. Strategically, to safeguard the decision area or region of interest (ROI), authors embedded EPR into the Region of Non-Interest (RONI). A similar effort was made in [24] where authors explored Pixel Value Differencing (PVD) so as to localize contrast regions which was then processed with Hamming code based data embedding as discussed [23].

Most of the approaches, particularly image transformation approaches have applied standard wavelet transform techniques to perform data embedding in image steganography. However, realizing the singularity problems, most of the approaches are found confined. To deal with this, authors [25] applied Ripplet-II transform by performing high dimensional generalization of the ridgelet transform where they incorporated a new variable called degree (d). However, they focussed on medical image classification and hence could not be explored for its efficacy to enable reversible steganography. However, they found that RT can perform better feature extraction or image transformation that could enable an array of enhanced computations. A similar outcome was realized by Dhaarani et al. [26] who examined its efficacy for grey scale medical image compression. However, they [26] obtained RT by generalizing the curvelet transform. To enhance performance, authors [26] applied RT to signify the images or 2D signals at distinct scales and orientations and applied Huffman coding to encode significant coefficients. Authors [26] found that RT exhibits satisfactory to achieve better compression ratio and low mean square error. In [27] RT technique (Ripplet Transform Type-I) was exploited significantly to enable multimodality Medical Image Fusion (MIF). Authors derived Ripplet Transform Type-I (RT) in conjunction with Pulse-Coupled Neural Network (PCNN). Authors found that Ripplet Transform can be a better alternative to perform image decomposition [26][27] that eventually could play vital role in medical image steganography.

Literature reveals that applying evolutionary computing techniques such as genetic algorithm (GA) can be effective for steganography, particularly for embedding capacity optimization and imperceptibility [28-30]. As empirical study, authors [29] applied GA for steganography encoding on the JPEG images for its security. GA was further applied in [30], where it was used for block mapping in LSB embedding. In [31] a GA based steganography was developed where authors applied discrete cosine transforms (GASDCT) and GA. Authors found that the use of GA enabled lower bit error rate. Applying standard DWT based image decomposition and GA based mapping, authors [32] developed a steganography model that embeds data into DWT coefficients, enriched with OPAP based optimization. Recently, authors [33] derived a steganography model for biomedical purpose. They applied queue data structure for communication where at first they encrypted secret message by means of Rabin Cryptosystem that eventually provided multiple blocks and sub-blocks to be distributed equally. In their approach, the receiver had four distinct values for plain text related to the single cipher text so that only authorized receiver could retrieve the original medical data. In [34] authors developed an encryption assisted with LSB embedding model for patient information hiding in medical image. A similar effort was made in [35], where authors applied AES encryption over secret information to be embedded into cover image. However, authors applied discrete cosine transform (DCT) to perform image decomposition and hiding. Further, in [36], authors exploited RSA and wavelet transform to perform Hash-LSB based data embedding steganography.

This is the matter of fact that a number of researches have been done to perform data embedding in images/medical image using steganography techniques; however most of the approaches use wavelet transform technique and are focussed on either PSNR enhancement or embedding capacity enhancement. However, the key requirements, such as ROI preservation, maximum imperceptibility, minimal or negligible histogram variations, statistical attack resilience, higher PSNR etc haven’t been considered. Considering these as motivations, in this paper a novel Ripplet Transform, LSB embedding, Adaptive GA based OPAP, RSA based security key encryption has been developed (dual level biomedical image security). The discussion of the proposed steganography model is given in the following sections.

## III. OUR CONTRIBUTION

With continued research and improvement in algorithm design, steganography can be taken as a serious means to hide data and the present work appears to be more efficient in hiding more data. Unlike traditional IWT based approaches, this paper applies Discrete Ripplet Transformation Type-I (DRT Type-I) -algorithm to perform image decomposition followed by secret message (i.e., patient information and previous diagnosis details, doctor details). Being a LSB embedding approach, we have applied an enhanced evolutionary computing approach called Adaptive Genetic Algorithm (AGA) to perform block mapping that enables not only the error reduction (between stego image and the cover image), but also preserves regional image properties by means of optimal pixel adjustment process. This process achieves higher embedding capacity and imperceptibility; thus, enabling an optimal solution for biomedical image communication over uncertain channels. Furthermore, the use of AGA based block flipping strengthens the statistical attack (for example, RS-attack) resiliency. To alleviate the issue of RS-attack on the stego image throughout channel, the impact of the inter-relation between pixels is compensated by using AGA that estimates the optimal adjusting mode for which the image imperceptibility could be maintained optimal. In our proposed AGA based block-mapping function, the, a chromosome is encoded as an array of 64 genes comprising permutations [1 to 64] that signifies number of pixels in each block. Thus, in each block, the data is embedded on multiple bites of the individual pixel. One of the key novelties of the proposed steganography model is the use of dual-security feature, where the secret information are at first encrypted with RSA cryptosystem, which is them followed by LSB embedding in the cover image. Here, in LSB embedding, our model embeds or hides information by replacing more than one bit from each pixel of the cover image. Being a reversible steganography technique, the developed model performs data embedding at the transmitter end, while extraction is performed at the receiver, where inverse DRT Type-I is applied to extract embedded secret information (i.e., patient details, diagnosis details etc). To examine the efficiency of the proposed steganography model biomedical data security for telemedicine purpose, the performance has been examined in terms of peak signal to noise ratio (PSNR), entropy, histogram deviations, R and S parameters etc.

## IV. SYSTEM DESIGN

The overall proposed biomedical image steganography (MIS) model comprises four phases:

1. Secret Patient Information Embedding,

2. GA Based Optimal Pixel Adjustment Process (OPAP),

3. Information Extraction,

4. RS Stegnalysis for DRT enriched Medical Image Steganography (MIS).

A brief discussion of the proposed research phases is presented as follows:

### *A. Secret Patient Information Embedding*

Since, the proposed steganography model contains patients information containing disease details, diagnosis made, doctor's details etc are in text which is further embedded to the biomedical diagnosis image (say, cover image); our proposed model can be stated as a multi-modal steganography. The overall proposed steganography model is presented in Fig. 1. As depicted in the following figure (Fig. 1), it can be found that the proposed system at fist accepts the color biomedical image as cover image, where the text information containing patient's details, diagnosis details etc are to be embedded. As stated, in this research an additional security feature has been proposed to enable a dual-security feature based biomedical data communication, the secret key information is at first processed using RSA cryptosystem algorithm. To ensure this additional security feature, the user defined security key has been considered which is supposed to be shared with the recipients to avoid decoding during transmission. Now, to perform secret data embedding, unlike traditional integer wavelet transform (IWT) algorithms, we have derived an efficient Discrete Ripplet Transformation (DRT) algorithm to decompose medical cover image into 8x8 blocks.

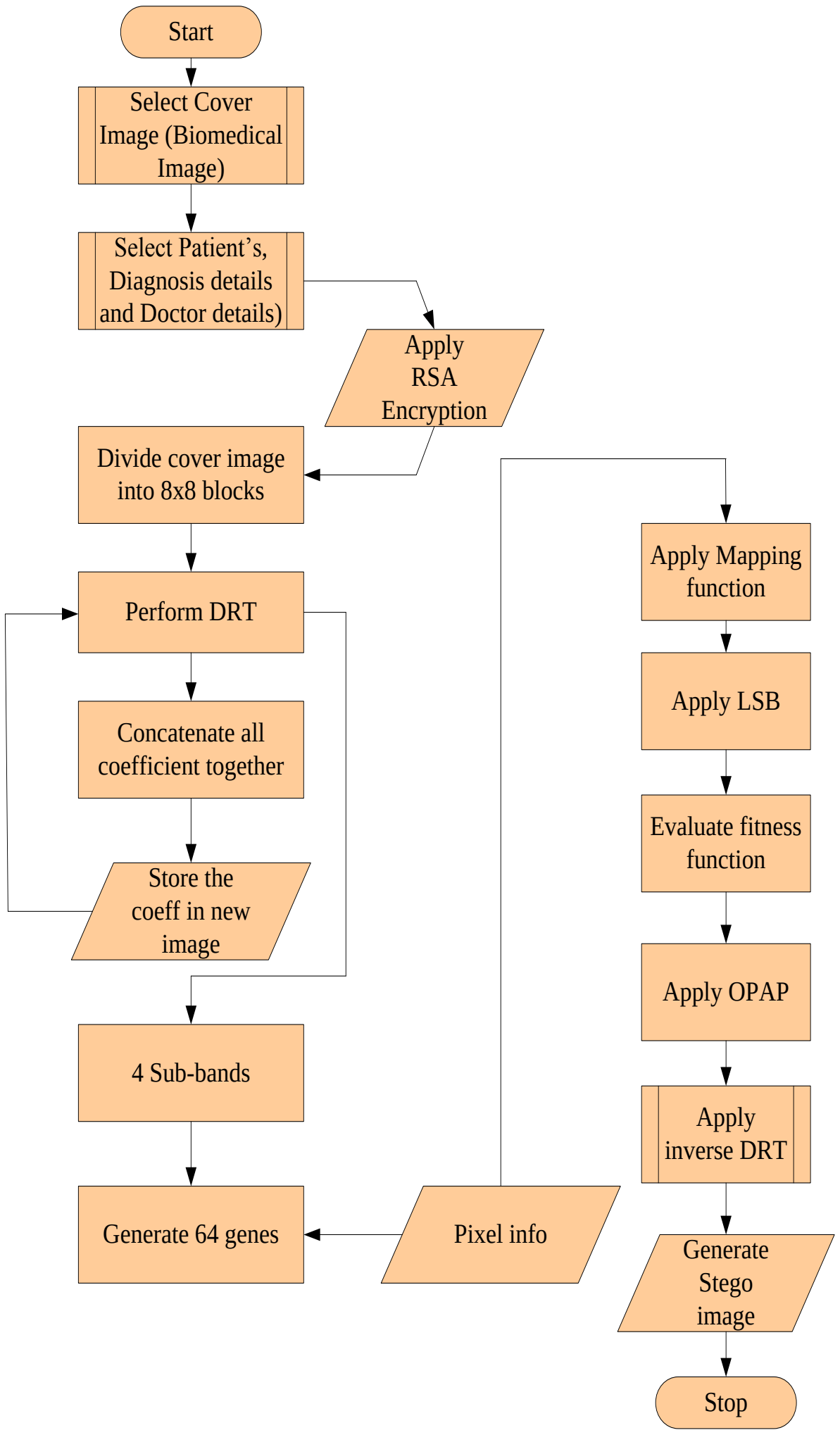

Fig. 1. Secret Patient's information embedding

Before discussing the message embedding, the following sub-section summarizes the discrete Ripplet transformation (DRT) technique and its efficacy over traditional wavelet transformation approaches to enable optimal data embedding and imperceptibility.

1) *Ripplet Transformation*

To achieve imperceptibility, maintaining efficient representation of the biomedical image is a must. In addition to, achieve an optimal steganography, maintaining higher image quality, embedding capacity (also called payload capacity), minimal computational complexity etc is the inevitable need. All these objectives can be achieved by means of certain efficient transformation technique. However, the generic transformation approaches such as Fourier transformation, wavelet transformation techniques do suffer from various issues such as discontinuities like edges in biomedical image. This as a result can degrade the significant image information. Thus, creating ambiguity towards effective diagnosis. To alleviate such limitations, in this paper, we have derived a new image decomposition model using Ripplet Transform (RT) which can be considered as a higher dimensional generalization of the curvelet transform, especially derived for presenting image at distinct scales and directions. It enables subjective support c and the degree d that establishes a dominator of standard curvelet transform. The acceptability of this model gets increased significantly when it has to be applied with biomedical processes, where there could be significant image information distributed across image. On the contrary, the other approaches such as Fourier transform technique can merely facilitate presentation of certain smooth images, is unable to preserve quality of image containing edges. Typically, edges introduce singularity problem or discontinuity in the intensity of the images. On contrary, presenting singularity effectively in medical image is a highly intricate problem, particularly for harmonic analysis. According to Gibbs phenomenon, the 1D singularity often leads destruction of the sparsity of the Fourier series presentation of a function. Interestingly, unlike Fourier transform, wavelet transform techniques can represent certain function with 1D singularity [37, 38]; however it is unable to deal with the issue of 2D singularity problem. This is because; standard 2D WT is a tensor product of 1D WTs that solves the 1D singularity in horizontal as well as vertical direction. To deal with such limitation, authors [39, 40] developed a ridgelet transform technique which is capable of resolving 1D singularities in any random direction. Being founded over Radon transform [41], Ridgelet transform provides significant information about orientation of the linear edges in the image that eventually enables extraction of the lines of arbitrary orientations. As ridgelet transform too was found to be confined to alleviate 2D singularity issue, authors [42-45] developed a curvelet transform technique that exhibited solution along smooth curves.

In general, Curvelet transform applies the concept of parabolic scaling so as to achieve anisotropic directionality that eventually gives hope to resolve 2D singularity problems along curves [46, 44, 45, 47]. However, the universal efficacy of parabolic scaling often remains questionable for different boundaries conditions, which is common in case of medical image data. In addition, the optimality of scaling concept is also not justified so far. Therefore, to deal with such issues, Ripplet transform has been derived that generalizes curvelet transform by incorporating two additional attributes, support c and degree d. These additional parameters, c and d facilitate RT with anisotropy ability to represent singularities along an arbitrary curve. Some of the key strengths of the RT approach are, multi-resolution, good localization due to frequency domain support and fast decay rate in spatial domain, hierarchical support for image presentation, better directionality, higher scalability, anisotropic efficiency and swift decay rate of the coefficients (that results into higher energy concentration). Considering biomedical image processing and key fundamental requirements, in this thesis RT technique has been applied to perform cover image decomposition and further coefficient embedding to hide secret information using LSB embedding approach.

As already stated, curvelet transform applies parabolic scaling concept to retrieve the properties of anisotropic directionality. This ability resolves singularity issue along $S^2$ curves. On the other hand, RT resolves singularity problem by presenting image with the discontinuities $S^d$. Here, we incorporate two parameters s, and d to generalize curvelet transform.

Mathematically, the derivation of the RT approaches is discussed as follows:

*2) Continuous Ripplet Transform (CRT)*

Typically, continuous ripplet transform (CRT) is stated as the inner product of 2D integral functions $C(\vec{x})$ and ripplets $P_{a\vec{b}c}(\vec{x})$, which is, mathematically, presented as (1).

$$R(a\vec{b}\theta) = \left(C, P_{a\vec{b}\theta}\right) = \int_0^{\infty} C(\vec{x})\overline{P_{a\vec{b}\theta}(\vec{x})d\,\vec{x}} \quad (1)$$

where $R\left(a\vec{b}\theta\right)$ signifies the RT coefficient and the other parameters, $a$, $\vec{b}$, and $\theta$ are the scale position, and the rotation parameters. Mathematically, the Ripplet in transform domain can be expressed as (2)

$$\hat{P}_a(r,w) = \frac{1}{\sqrt{S}} a^{\frac{1+d}{2d}} U(a,r) V\left(\frac{a^{1/d}}{S.d}.\omega\right) \quad (2)$$

where $\hat{P}_a(r,w)$ refers the FT of ripplet, and r and ω signify the radial and angular parameters, respectively. The other parameters, $U(r)$ and $V(\omega)$ signifies the radial and the angular windows, respectively. The introduced two variables d and S signify the degree and the support, correspondingly. To perform inverse CRT, the following expression is used:

$$C(\vec{x}) = \int R\left(a\vec{b}\theta\right) P_{a\vec{b}\theta}(\vec{x}) dad\,\vec{b}\ {}^{d\theta}\!/_{a^3} \quad (3)$$

*3) Discrete Ripplet Transform(DRT)*

Unlike CRT, the discrete Ripplet transform (DRT) is performed by discretizing the RT coefficient$(a, \vec{b}, \theta)$. In DRT, scale parameter a is discretised at dyadic intervals while other remaining parameters $\vec{b}$ and θ are discretised at equal-spaced difference or intervals. These parameters are replaced with ($a_x$, $\vec{b}_y$ and $\theta_z$) respectively and thus, the following are obtained:

$$a_x = 2^{-x} \quad (4)$$

$$\vec{b}_y \quad (5)$$

$$\theta_z \quad (6)$$

where $y = [y_1 y_2]^T$ and $(x, y_1, y_2, z)\varepsilon k$.

The frequency response of the RT (in transform domain) is obtained using (7):

$$\hat{P}_a(r,w) = \frac{1}{\sqrt{S}} a^{\left(\frac{1+d}{2d}\right)} U(2^{-x}.\gamma) V\left(\frac{2^{-x\left(\frac{1}{d}-1\right)}}{S}.\omega - 1\right) \quad (7)$$

The DRT of the 2D image (signal) $C(i,j)$ having dimension $M \times N$ is usually presented in terms of the RT coefficients, given by $R_{x\overline{y}\overline{z}}$. Mathematically,

$$R_{x\overline{y}\overline{z}} \quad (8)$$

Finally, the image $\hat{C}(i,j)$ is reconstructed by applying inverse DRT given as (9).

$$\hat{C}(i,j) = \sum_{x=0}^{1}\sum_{\vec{y}=0}^{1}\sum_{z=0}^{1} R_{x\vec{y}z}(i,j) \quad (9)$$

Now, considering our proposed steganography and associated secret information hiding mechanism, at first the DRT algorithm is applied over the cover image (i.e., biomedical image) and the ripplet coefficients are obtained, which is then followed by secret information hiding using LSB embedding approach. The proposed secret embedding approach is presented as follows:

*4) DCT Based Secret Message embedding*

The sequential implementation of the DCT based secret message embedding is presented as follows:

**Step-1** Obtain the Biomedical Image data and apply it as the cover image $C(x,y)$.

**Step-2** Obtain the patient's information, diagnosis, and doctor details as secret data $S(x,y)$.

**Step-3** Implement histogram modification as pre-processing over biomedical image (or cover image).

**Step-4** Perform decomposition of the pre-processed cover image using DRT and obtain the DRT coefficients $I$ using (10).

$$I = DRTcoeff_{R(a\vec{b}\theta)}\left(C(x,y)\right) \quad (10)$$

**Step-5** To apply least significant bit (LSB) embedding to hide the secret patient details or the secret information.

In this process, the DRT coefficients are selected by sorting them in increasing order. Thus the significant DRT coefficients are retrieved. Interestingly, with the sorting approach, it raises the bit-ordering issue that as a result introduces un-embedding capacity and hence degrades payload capacity. In this research work, to alleviate such issue, a novel enhanced evolutionary computing (EC) algorithm named Adaptive Generic Algorithm has been developed that, eventually retrieves the optimal scrambled DRT coefficient position at each block. In other words, the proposed AGA model performs block flipping so as to obtain the optimal coefficient position where the data is hidden.

**Step-6** Once performing scrambling of the DRT coefficient, the orientation (θ) having the maximum energy (or) the highest variance at the individual block (M × M) are considered as the significant coefficient to embed or hide the secret information. Mathematically,

$$C_m(x,y) = \max_{v}\left(var\left(DRTCoeff_{C_{m(x,y)}}(1{:}m + 1, v)\right)\right) \quad (11)$$

**Step-7** Now, the retrieved DRT coefficients are updated to a matrix $R_{\mathrm{m}}$ for m[th] block.

$$R_{\mathrm{m}} = \begin{bmatrix} DRTCoeff_{C_{m(x,y)}} & [1, C_m\,(x,y)], \\ DRTCoeff_{C_{m(x,y)}} & [2, C_m\,(x,y)], \ldots. \\ DRTCoeff_{C_{m(x,y)}} & [m+1, C_m\,(x,y)] \end{bmatrix}^{\mathrm{T}} \leq \mathrm{m} \leq \mathrm{M} \quad (12)$$

In addition, to avoid perceptibility problem and statistical attack issues such as RS-Attack, we have incorporated a novel AGA based OPAP optimization model that estimates the optimal pixel differences so as to enable higher message or payload embedding capacity and low entropy.

**Step-8** Now, obtaining the coefficient matrix the column elements are put into the matrix $R$.

$$R = [R_1, R_2, R_3 \ldots\ldots.. R_j] = \{R_{xy}\}\ with\ x = 1,2,3 \ldots i\ \&\ y = 1,2,3 \ldots\ldots \quad (13)$$

where, i, j variables signify row and column of the cover image or the biomedical image.

**Step-9** Now, the patient's diagnosis details or the consi converted to secret data that appends supplementary proposed dual security based steganography.

**Step-10** Thus, the secret data $S = \{S_1, S_2 \ldots\ldots S_j\}$ is emb

$$R^S(x,y) = R_j + \eta S R_j$$

where, $\eta$ signifies the scaling factor which we have selected as $\eta = 0.30$. Thus, with embedded data the stego image is formed which is transmitted over network.

**Step-11** Now, perform optimal pixel placement algorithm to avoid any statistical attack such as RS attack and to strengthen imperceptibility. The discussion of the proposed RS-Resilient block flipping and OPAP is discussed in the next section.

Once receiving the stego image at the receiver end, the inverse DRT (IDRT) is applied over the stego image to retrieve cover image. In addition, being RSA encrypted, our proposed model requires recipient to use the same key applied for encryption, so as to retrieve the secret information for decision process.

*B. Genetic Algorithm Based Optimal Pixel Adjustment Process (OPAP)*

The predominant objective of OPAP is to reduce the error in between the stego image and the cover image (i.e., medical image). For illustration, in case the pixel index of the cover image is 10000 (signifying 16 as the decimal number) and the secret data vector to be embedded for 4 bits be 1111, then it would result into the change in the pixel index and would become 11111 (characterizing 31 as digital number). In this case, the embedding error would be 15 while, once implementing OPAP the fifth bit is targeted to be made 0 (from 1 to 0). Thus, the error would be minimized to 1. The overall OPAP model can be characterized into the following approach:

Case 1 $(2\mathrm{k}-1 < \delta i < 2k)$: if $\mathrm{pi}' \geq 2\mathrm{k}$, then $\mathrm{pi}'' = \mathrm{pi}' - 2\mathrm{k}$ otherwise $\mathrm{pi}'' = \mathrm{pi}'$;
Case 2 $(-2\mathrm{k}-1 < \delta i < 2k-1)$: $\mathrm{pi}'' = \mathrm{pi}'$;
Case 3 $(-2\mathrm{k} < \delta\mathrm{i} < -2\mathrm{k}-1)$: if $\mathrm{pi}' < 256 - 2\mathrm{k}$, then $\mathrm{pi}'' = \mathrm{pi}' + 2\mathrm{k}$; otherwise $\mathrm{pi}'' = \mathrm{pi}'$;

Here, the variables $P_i, P_i^{ó}$ and $P_i^{ó}$ represents the associated pixel values of the ith pixel in the three images; cover image, stego image and the retrieved image by the generic LSB embedding process, respectively.

In above expression, the variable $\delta_i(= P_i^{ó} - P_i)$ signifies the embedding error in between the stego image and the cover image [6]. Thus, once embedding the n-LSBs of the cover image $P_i$ with n message bits, $\delta_i$ would be (15):

$$-2^n < \delta_i < 2^n \tag{15}$$

In this paper, Adaptive Genetic Algorithm, an enhanced evolutionary algorithm with adaptive GA parameter update (crossover probability and mutation probability) capability, has been applied to perform OPAP. Here AGA obtained the mapping function for each image blocks. In our proposed OPAP model, a chromosome is encoded as an array of 64 genes possessing the permutations 1 to 64 that signify indicate the number of pixel in individual block. GA obtains the best adjustment matrix. In the proposed work, PSNR is considered as the fitness value of the GA based optimization, where after each iteration or generation we expect to reduce the PSNR (16).

$$\mathrm{PSNR} = 10\ \log_{10}\frac{\mathrm{M}\times\mathrm{N}\times 255^2}{\sum_{i,j}\left(y_{i,j}-x_{i,j}\right)^2} \tag{16}$$

where M and N represent the image sizes, while x and y signify the intensity before and after message embedding.

*C. Information Extraction*

In our proposed model, once receiving the stego image at the recipient terminal, at first it is processed for inverse DRT where the LSB coefficient from DRT coefficient (in coefficient domain with coefficients with each $8 \times 8$ blocks) is selected, which is then placed in a matrix $R_m^*$. It is then processed for descrambling so as to substitute the coefficient in its own-place and the pixel sequence is obtained. In this model, we estimate a correlation estimator $\mathbb{W}$, which signifies the mean correlation in between the individual row off $R^*$ is obtained by (17)

$$\mathbb{W} \tag{17}$$

Thus, the patient data transmitted through medical image is obtained by equation (18) and the medical data transmitted is reconstructed by means of the inverse DRT.

$$S_y^* = WS_y \tag{18}$$

In addition, to retrieve the accurate message data or the patients' information, diagnosis and doctor details are obtained by applying decryption key which is used in the same for encryption at the transmitter terminal.

The overall process of the information extraction is presented in Fig. 2.

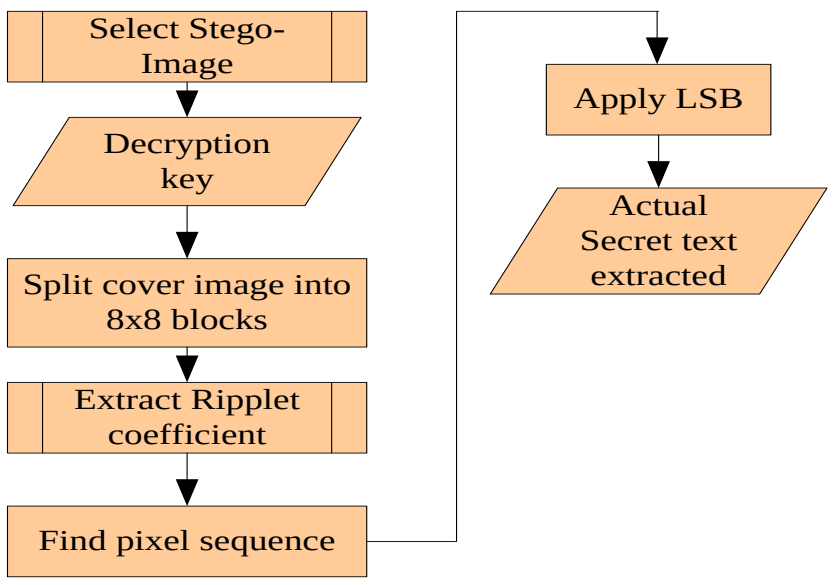

Fig. 2. Patient information (secret message) extraction

### *D. RS Stegnalysis Enriched OPAP for Medical Image Steganography*

Undeniably, there has been different type of attacks functional on the basis of statistical correlation between pixels. On the other hand, realizing the fact that embedding text information or the patient's information inside cover image could lead entropy. Thus, making image perceptible. To avoid this issue, in this paper, we have enriched our proposed medical image steganography technique to deal with RS type attacks in the channel. This section discusses the RS Stegnalysis and RS resilient MIS and a novel AGA based RS resilient MIS model for secure medical data communication over uncertain channels.

In the process of RS steganalysis, three distinct types of block flipping named, positive ($F_P$), negative ($F_N$) and Zero-flipping ($F_0$)are performed. The positive flipping $F_P$ refers the transformational inter-relation between 2i and $2i + 1$ pixel, which is equivalent, to LSB. On the other hand, the negative flipping $F_N$ refers the transformational inter-relation between $2i - 1$ and2i. Thus, the inter-relation between these two distinct flipping can be derived as follows:

$$F_N = F_P(x + 1) - 1 \tag{19}$$

Here, an identity permutation is defined in terms of Zero-flipping.

$$F_0(x) = x \tag{20}$$

Let, $F_0$, $F_1$ and $F_{-1}$be the flipping functions. The flipped group is obtained by means of executing these flipping functions on each pixel of the medical image block. Mathematically, the flipping group $F(G)$ is presented as

$$F(G) = (F_{M(1)}(x_1), F_{M(2)}(x_2), \dots, F_{M(n)}( \tag{21}$$

In (21), the variable $M = M(1), M(2), \dots, M(n)$ signifies a flipping mask, where $M(i) \in \{1, 0, -1\}$.

The parameter G is regular in case of $f(F(G)) > f(G)$, and G is singular if and only if $f(F(G)) < f(G)$.

To perform RS Stegnalysis, at first, the medical image is split into multiple non-overlapping blocks, where the individual is re-arranged to form a vector $G = (x_1, x_2, \dots, x_n)$ in certain arbitrary order. Here, the inter-relation or the correlation between pixels can be estimated using equation (22).

$$f(x_1, x_2, \dots, x_n \sum_{i=1}^{n+1} |x_i - x_{i+1}| \tag{22}$$

where x refers to the pixel value, while n presents the total number of pixels.

On the other hand, f signifies the spatial correlation in between the adjoining pixels. Here, smaller f refers stronger to correlation. Once estimating $f(G)$, non-negative flipping and non-positive flipping are applied on the individual block. It is then followed by the use of (22) to estimate the $f(F(G))$ in individual block. Let $R_m$ and $S_m$be the relative number of the regular blocks and singular blocks after positive flipping. Similarly, $R_{-m}$ and $S_{-m}$ be the relative number of regular and singular blocks after the negative flipping. It is found that there exists a relationship between these blocks in the following manner:

$$R_m \approx R_{-m}, S_m \approx S_{-m} \text{and } R_m > S_m, R_{-m} > S_{-m}$$

Interestingly, the difference usually increases between the relative number of the regular blocks $R_m$ and $R_{-m}$increases as per the size of the message embedded into the image. Similar feature is observed for the difference between singular blocks $S_m$ and $S_{-m}$.

#### *1) RS Steganalysis the Resilient Medical Image Steganography*

On the basis of aforementioned logical discussion, in the proposed AGA based RS Stegnalysis model the pixel values are adjusted so as to maintainRm $\approx S_m, R_{-m} \approx S_{-m}$. Since, embedding patient's details or secret message information into the biomedical image data alters the bits in higher planes and that consequently degrades the imperceptibility of the stego image, merely the 2nd and 3rd lowest bit-planes are altered. For illustration, let B be the initial value of an image block and considering the alteration in the second lowest bit-plane, the variations incorporated between the initial (i.e., original image block) and altered block can be visualised as an adjustment matrix such as $A_1$ or $A_2$. Let the altered image blocks be $B'_1 = B + A_1$ and $B'_2 = B + A_2$. For illustration, for the initial or the original block $B, f(B) = 99$ and $f(F_{_}(B)) = 120$, where $F_{_}$ signifies the non-positive flipping. Now, for altered block $B'_1, f(F_{_}(B'_1)) = 90$, only when F is non-positive flipping. Similarly, in case of other altered block $B'_2, f(F - (B'_2)) = 150$. These all illustrations state that the kind of the block (regular or singular) can be easily changed by means of certain effective adjustment. This as a result can ensure that RS steganalysis could not detect the presence of certain hidden message or text within the stego image.

To deal with the RS attack in channel, in this paper, we have derived proposed steganography model as RS resilient model, where AGA has been applied to estimate the optimal adjustment matrix. AGA transforms an optimization in the form of chromosome evolution phenomenon. Once retrieving the optimal solution (i.e., individual chromosome with expected fitness value), it gives optimum or sub-optimum solution. Similar to the above mentioned adaptive genetic algorithm, to incorporate RS resiliency, we have varied crossover and mutation probability adaptively where these value varies as per the chromosomes having similar fitness value. It reduces not only the computational overhead and time, but also avoids local minima and convergence type issues which is common in major standard learning approaches including GA and neural networks. In this model, once embedding the patients information inside the DRT coefficients using LSB embedding approach, AGA has been applied to perform adjustment (say enhanced OPAP).

In this process, the stego image is split into $8 \times 8$ blocks and the obtained blocks are labelled as follows:

1. For block B, implement non-positive and non-negative flipping on the block and generate the flipping mask $M+$ and $M-$, randomly to obtain $B'_+$ and $B'_-$, respectively.
2. Estimate $f(B'_+), f(B'_-)$ and $f(B)$.
3. Declare four variables to classify blocks by comparing$f(B'_+), f(B'_-)$ and $f(B)$.
   - $P_{+R}$, event count when the block is regular under non-negative flipping.
   - $P_{+s}$, event count when the block is singular under non-negative flipping
   - $P_{-R}$, event count when the block is regular under non-positive flipping
   - $P_{-s}$, event count when the block is Singular under non-negative flipping
4. Perform comparison between $P_{+R}$ and $P_{+s}$, and $P_{-R}$ and $P_{-s}$, and estimate the block labels:
   - $R+, if\ P_{+R}/P_{+s} > Th.$
   - $S+, if\ P_{+s}/P_{+R} > Th.$
   - $R-, if\ P_{-R}/P_{-s} > Th.$
   - $S-, if\ P_{-s}/P_{-R} > Th.$

In our model, the threshold value $Th$ is taken as 1.8.

5. Thus, on the basis of block labels these are classified into 4 distinct groups $R+R-, R+S-, S+R-$, and $S+S-$.

In our proposed model, the remaining blocks (apart from the four categories as defined above), are excluded from further processing.

In comparison to the original medical image, the value of $R+R-$ and $S+R-$ blocks usually increases in the stego images after message embedding. In typical RS steganalysis model or RS analysis, this resulting feature is detected that signifies the presence of certain hidden data in the image. Therefore, to alleviate this issue, in this paper, we intend to reduce the value of $R-$ blocks and to achieve this, we have applied AGA algorithm that employs the following steps.

1. **Chromosome Initialization:** Starting from the first pixel in each block we select each three adjoining pixels as the initial chromosome.
2. **Reproduction and Mutation:** In this process, the second lowest bits in the chromosomes are flipped arbitrarily that as a result generates numerous 2nd generation chromosomes $C_i$.
3. **Selection: In selection process,** the chromosome having best fitness is selected to substitute its subsequent initial chromosome. In our work, we have applied the following equation (23) to perform fitness estimation.

$$Fitness = \alpha(e_1 + e_2) + PSNR \quad \alpha \ldots weight \tag{23}$$

where, $e_1$ refers likelihood of $f(F_{_}(C_i)) < f(C_i)$ and $e_2$ signifies the likelihood of $f(F_+(C_i)) > f(C_i)$. The parameters, α presents the weight estimated by experimental process, which is applied to control the visual quality of the stego image after message embedding. In addition, it is used to control the weight for hidden data security provision. For illustration, for a pre-defined α value, higher value of $e_1$ and $e_2$ signifies higher security provision, which is significant for biomedical data transmission over insecure or uncertain channel conditions. With this motivation, in our model we intend to enhance the value of fitness function, i.e., PSNR.

4. Estimate the adjusted image blocks $P_{-R}$ and $P_{-S}$. The condition when $P_{-S} > P_{-R}$, the block is considered as optimally adjusted.

5. Crossover. To perform crossover, the chromosomes are shifted by one pixel, which is then followed by Step 2. Here, it should be noted that based on the fitness value in our approach, the number of pixels to be shifted does vary. Thus, enabling adaptive genetic algorithm (AGA) implementation.

Once performing block adjustment, we have calculated the value of $R_m$ , $R_{-m}$, $S_m$ and $S_{-m}$ of the biomedical image under study and in case the difference between $R_m$ and $R_{-m}$ (and/or$S_m$ and $S_{-m}$) is more than 5%, the next block is adjusted. This overall process not only enhances the embedding capacity but also strengthens the imperceptibility of the stego image.

The results obtained in this study are discussed in the next section.

## V. Results and Discussion

This paper has presented a novel and robust evolutionary computing enriched RS attack resilient medical image steganography (MIS) model to be used for telemedicine purposes. Unlike traditional approaches, our proposed method can be considered as multi phase optimizations model to achieve a robust steganography, where at first to ensure optimal security a dual security feature has been incorporated. To achieve this at first a generic RSA cryptosystem was applied over patient's details and corresponding diagnosis or doctor information. On the other hand, to strengthen steganography model a robust Discrete Ripplet Transform (DRT) technique was derived so as to perform message embedding in transform domain. Unlike existing integer wavelet transformation (IWT) techniques, our proposed DRT method alleviates major singularity problem. In addition it assures edge information preserving, which is of a paramount significance for medical data to enable flawless diagnosis decision in telemedicine. Thus, applying DRT, the LSB embedding has been performed so as to hide secret data into medical image. Further, considering the inevitable need to enrich imperceptibility and higher embedding capacity, in this research work an efficient evolutionary computing algorithm named Adaptive Genetic Algorithm (AGA) based optimal pixel adjustment process (OPAP) has been developed with intention to reduce the entropy introduced due to message embedding. Now, realizing the fact that message embedding in medical image do often alters the pixel uniformity and hence applying any statistical relationship between pixel the presence of any secret image could be traced. RS attack is dominating one statistical analysis approaches to trace hidden secret information in image. Considering this as a motivation, in this research paper, we intended to derive a novel RS resilient steganography model, AGA based block flipping approach has been applied over stego image (i.e., biomedical image with secret information). The performance of the overall proposed steganography model is examined in terms of peak signal to noise ratio (PSNR), embedding capacity, RS analysis or steganalysis results before and after AGA based flipping and optimization. To assess the performance of the proposed model, the investigation has been made for both the color magnetic resonance image (MRI) images as well as gray MRI images. Similarly, varying the payload, the PSNR and histogram variations have been analyzed. Being a reversible image steganography model, inverse DRT has been applied at the receiver end to retrieve the embedded message from the stego image.

The overall model has been developed using MATLAB 2015a software tool, where a well developed Graphical User Interface (GUI) has been developed (Fig. 3) to feed the data (medical image or the cover data and patient information). The developed GUI facilitates user to select input data, and user defined key series to perform encrypt secret information at the transmitter end, before embedding it to the cover image. Further at the receiver to perform extraction, encryption key is used to retrieve the secret key; however before it Inverse DRT s applied to decompose stego image and retrieve hidden information from coefficients. The schematic and overall implementation from is presented in the following GUI (Fig. 3).

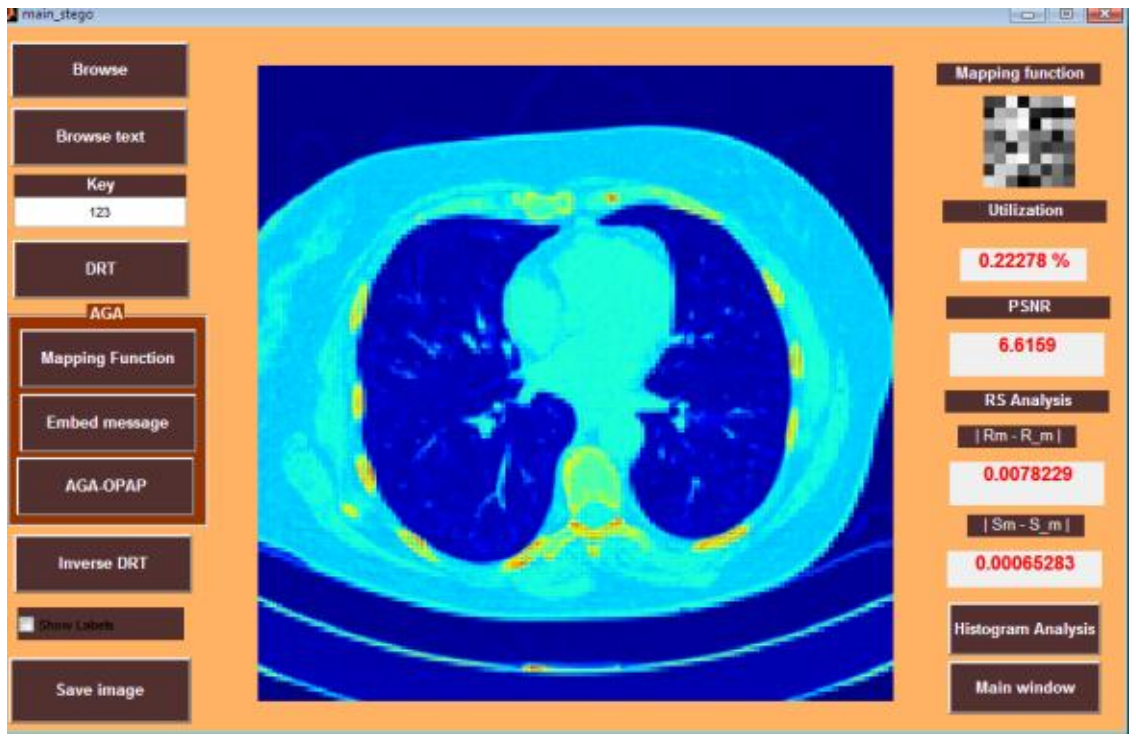

Fig. 3. Graphical User Interface to perform

Some of the data used for the experimental process is given in Table 1. Here, to assess efficacy of the proposed model both the color as well as gray biomedical image have been taken into consideration.

TABLE I. BIOMEDICAL DATASET

| | | | | |
|---|---|---|---|---|
| **Colour Biomedical Image** | 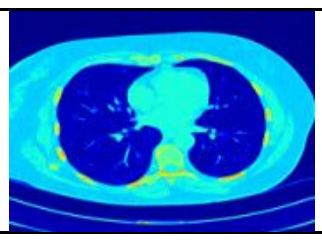 | 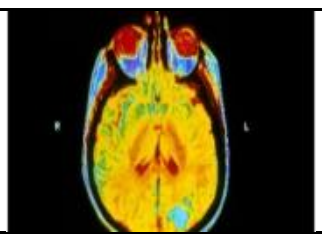 | 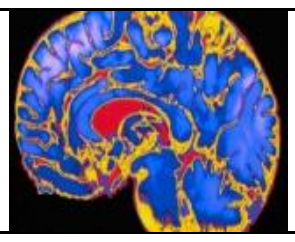 | 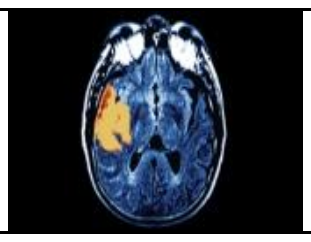 |
| | CBD1 | CBD2 | CBD3 | CBD4 |
| **Gray Biomedical Image** | 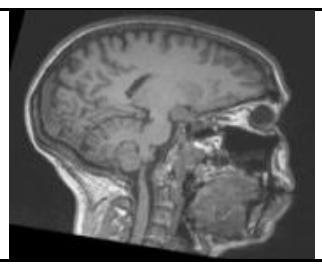 | 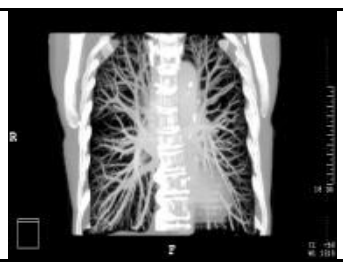 | 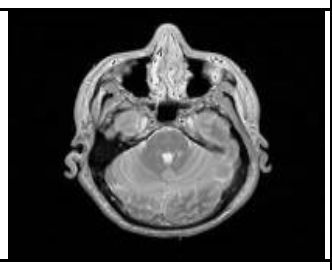 | 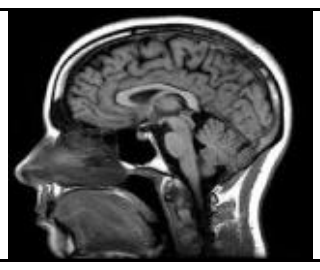 |
| | GBD1 | GBD2 | GBD3 | GBD4 |
| **Colour Image** | 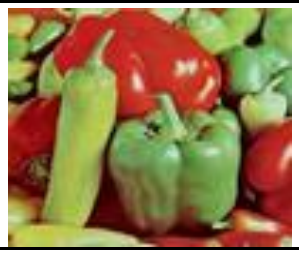 | 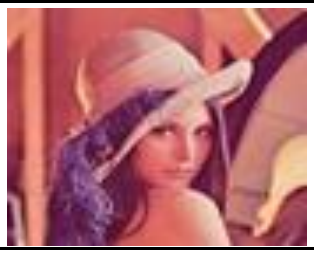 | 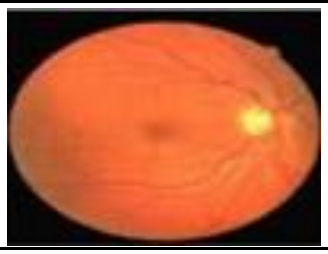 | 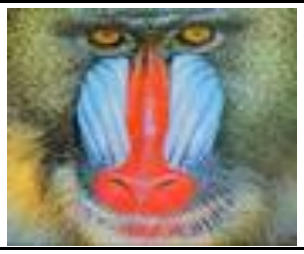 |
| | Pepper | Lena | Retina | Baboon |
| **Gray Image** | 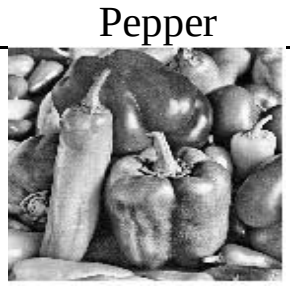 | 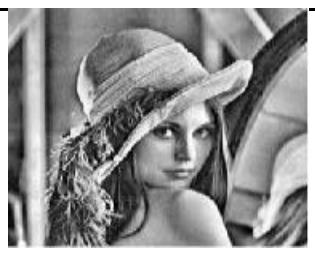 | 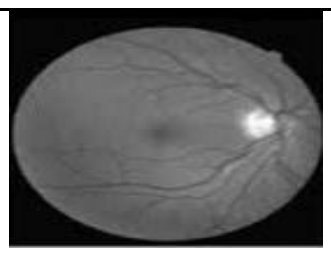 | 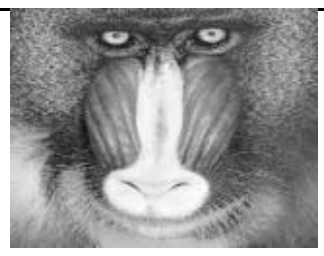 |
| | Pepper | Lena | Retina | Baboon |

At first the proposed steganography model has been assessed in terms of PSNR, where to examine the relative efficacy a parallel model using Integer Wavelet Transform has been developed. The algorithm has been developed for both the IWT as well as proposed DRT based message embedding.

Data used for the experimental process

TABLE II. PSNR ANALYSIS

| **data** | **IWT (Original Cover Image)** | **After OPAP** | **DRT (Original Cover Image)** | **After OPAP** |
|---|---|---|---|---|
| CBD1 | 26.048 | 26.615 | 37.028 | 39.664 |
| CBD2 | 31.61 | 34.71 | 41.54 | 44.00 |
| | 5 | 4 | 0 | 2 |
| CBD3 | 27.616 | 28.044 | 42.017 | 45.133 |
| CBD4 | 31.770 | 32.168 | 38.807 | 44.698 |
| GBD1 | 49.242 | 49.550 | 52.931 | 55.063 |
| GBD2 | 37.502 | 38.091 | 44.893 | 47.017 |
| GBD3 | 40.621 | 41.142 | 46.888 | 49.094 |
| GBD4 | 40.169 | 40.256 | 47.001 | 49.785 |
| C Pepper | 27.356 | 25.578 | 37.024 | 38.743 |
| C Lenna | 33692 | 34.964 | 41.511 | 42.532 |
| C Retina | 18.543 | 27.022 | 43.016 | 44.163 |
| C Baboon | 29.780 | 37.155 | 38.888 | 43.611 |
| G Pepper | 49.230 | 50.570 | 52.441 | 53.985 |
| G Lenna | 38.522 | 39.011 | 44.663 | 48.246 |
| G Retina | 41.671 | 43.134 | 46.808 | 51.298 |
| G Baboon | 40.169 | 43.256 | 48.901 | 52.752 |

As depicted in the above table (Table 1), the proposed discrete Ripplet transform based steganography enables higher PSNR as compared to the other traditional integer wavelet transformation based embedding technique. The results also reveal that the use of AGA based optimal pixel adjustment process strengthens the process of LSB embedding that as a result not only enhances the message bit placement uniformly across the cover image, but also its efficient extraction. Since, in our work, we have incorporated AGA for OPAP in both the cases, i.e., with IWT based embedding as well as DRT based patient information embedding; however the better performance by DRT based approach reveals that the image decomposition and information hiding at the coefficients is better with DRT technique. This is the fact that PSNR is greatly influenced at the receiver terminal where the medical image is extracted or decomposed to retrieve the hidden information, the consideration of pixel uniformity and sequential information retrieval is better with DRT technique that as a result enables better message retrieval and hence higher PSNR. In addition, the results also state that block based data hiding is better and computationally efficient than the pixel replacement based approaches. It should be noted that in our proposed DRT based steganography, block based data embedding has been performed,

TABLE III. EMBEDDING CAPACITY

| **data** | **Payload (bits)** | **PSNR (After OPAP)** *IWT based embedding* | *DRT based embedding* | **DRT based steganography Embedding Capacity (%)** |
|---|---|---|---|---|
| CBD1 | 4412 | 26.615 | 39.664 | 42.200 |
| | 6096 | 26.010 | 39.900 | 47.100 |
| | 6883 | 26.091 | 38.997 | 41.000 |
| CBD2 | 4412 | 34.714 | 44.715 | 42.000 |
| | 6096 | 33.071 | 44.610 | 44.010 |
| | 6883 | 33.004 | 44.511 | 45.170 |
| CBD3 | 4412 | 28.044 | 45.133 | 52.030 |
| | 6096 | 28.019 | 45.280 | 52.960 |
| | 6883 | 28.003 | 45.100 | 53.490 |
| CBD4 | 4412 | 32.168 | 44.698 | 33.180 |
| | 6096 | 32.110 | 44.550 | 34.320 |
| | 68 | 31 | 44 | 34 |
| | 83 | .1044 | .5125 | .6304 |
| GBD1 | 4412 | 9.5504 | 5.0635 | 2.0104 |
| | 6096 | 8.0214 | 5.0085 | 2.6504 |
| | 6883 | 6.0193 | 4.8904 | 2.9303 |
| GBD2 | 4412 | 8.0913 | 7.0174 | 3.1803 |
| | 6096 | 7.9143 | 5.0124 | 5.0103 |
| | 6883 | 6.1984 | 5.0074 | 6.6704 |
| GBD3 | 4412 | 1.1424 | 9.0944 | 4.0004 |
| | 6096 | 0.0183 | 9.0474 | 5.0104 |
| | 6883 | 7.1134 | 8.5304 | 7.0303 |
| GBD4 | 4412 | 0.2563 | 9.7854 | 7.9704 |
| | 60 | 7 | 9 | 5 |
| | 96 | .0713 | .4134 | .4104 |
| | 6883 | 3.8252 | 9.0024 | 9.0304 |
| C Pepper | 4531 | 7.1472 | 0.4174 | 5.9975 |
| | 6215 | 6.5302 | 0.6583 | 1.3374 |
| | 7002 | 6.6123 | 9.7374 | 4.6974 |
| C Lena | 4531 | 5.4083 | 5.5644 | 5.7004 |
| | 6215 | 3.7323 | 5.4574 | 7.9704 |
| | 7002 | 3.6642 | 5.3564 | 9.2355 |
| C Retina | 4531 | 8.6042 | 5.9904 | 6.7125 |
| | 6215 | 8.5792 | 6.1404 | 7.7265 |
| | 7002 | 8.5633 | 5.9564 | 8.3043 |
| C | 45 | 2 | 5 | 6 |
| Baboon | 31 | .8113 | .5474 | .1663 |
| | 6215 | 2.7523 | 5.3964 | 7.4083 |
| | 7002 | 1.7265 | 5.3575 | 7.7464 |
| | 4531 | 0.5404 | 6.1095 | 5.7904 |
| G Pepper | 6215 | 8.9814 | 6.0535 | 6.4884 |
| | 7002 | 6.9393 | 5.9324 | 6.7933 |
| | 4531 | 8.8523 | 7.9104 | 6.1663 |
| G Lena | 6215 | 8.6723 | 5.8674 | 8.1603 |
| | 7002 | 6.9214 | 5.8625 | 9.9704 |
| | 4531 | 1.9644 | 0.0264 | 7.9604 |
| G Retina | 6215 | 0.8183 | 9.9784 | 9.0605 |
| | 70 | 7 | 9 | 1 |
| | 02 | .8554 | .4525 | .2624 |
| | 4531 | 1.0613 | 0.7305 | 1.3874 |
| G Baboon | 6215 | 7.8123 | 0.3514 | 9.4965 |
| | 7002 | 4.501 | 9.933 | 3.442 |

Thus, observing above mentioned results (Table 2), it can be found that the proposed DRT based steganography outperforms IWT based approach in terms of higher embedding capacity as well as PSNR, even for higher payload. One more interesting outcome has been observed that when varying (precisely increasing) payload, the PSNR decreases in case of IWT based steganography. On the contrary, with DRT the decrease in PSNR is very small and can be percept as having negligible or very small affect on imperceptibility. The results affirm that the proposed steganography model performs uniform even varying payload without any significant PSNR degradation. Considering the effectiveness of the proposed DRT and AGA based steganography model, we have further enriched it with RS resiliency ability, where AGA has been applied to perform block flipping and block adjustment. Here, we considered the hypothesis that block adjustment can perform better than the standard pixel adjustment based OPAP. The following results present the RS analysis outcomes from the proposed steganography model.

In our model, we have applied Stegnalysis on all eight biomedical images. The results obtained are given In Table 3. Here, it can be found that the value of $R_m - R_{-m}$ and $S_m - S_{-m}$ increases as per increase in embedding capacity. It signifies the rise in the complexity to adjust the blocks so as to preserve the imperceptibility.

TABLE IV. RS ANALYSIS (COVER IMAGE)

| Data | $R_m - R_{-m}$ | $S_m - S_{-m}$ |
|---|---|---|
| CBD1 | 0.0093 | 0.0082 |
| CBD2 | 0.0052 | 0.0018 |
| CBD3 | 0.0010 | 0.0011 |
| CBD4 | 0.0026 | 0.0047 |
| GBD1 | 0.0003 | 0.0043 |
| GBD2 | 0.0011 | 0.0003 |
| GBD3 | 0.0004 | 0.0013 |
| GBD4 | 0.0044 | 0.0098 |
| C Pepper | 0.0099 | 0.0085 |
| C Lena | 0.0051 | 0.0015 |
| C Retina | 0.0012 | 0.0013 |
| C Baboon | 0.0025 | 0.0041 |
| G Pepper | 0.0003 | 0.0043 |
| G Lena | 0.0012 | 0.0001 |
| G Retina | 0.0004 | 0.0013 |
| G Baboon | 0.0044 | 0.0092 |

On the other hand, Table 5 presents the differences ($R_m - R_{-m}$ and $S_m - S_{-m}$), where it can be found that the value of R − decreases by using AGA based block mapping and block adjustment process that makes it difficult or even infeasible for attackers to identify or detect the secret information hidden in the cover image (biomedical image).

TABLE V. RS ANALYSIS ( STEGO IMAGE)

| data | $R_m - R_{-m}$ | $S_m - S_{-m}$ |
|---|---|---|
| CBD1 | 0.0078 | 0.0006 |
| CBD2 | 0.0024 | 0.0058 |
| CBD3 | 0.0020 | 0.0047 |
| CBD4 | 0.0031 | 0.0013 |
| GBD1 | 0.0097 | 0.0044 |
| GBD2 | 0.0052 | 0.0089 |
| GBD3 | 0.0097 | 0.0084 |
| GBD4 | 0.0043 | 0.0022 |
| C Pepper | 0.0078 | 0.0006 |
| C Lena | 0.0024 | 0.0058 |
| C Retina | 0.0020 | 0.0047 |
| C Baboon | 0.0030 | 0.0013 |
| G Pepper | 0.00979 | 0.0041 |
| G Lena | 0.0051 | 0.0082 |
| G Retina | 0.0097 | 0.0082 |
| G Baboon | 0.0043 | 0.0020 |

Thus, the overall results affirm that the proposed steganography approach achieves optimal balance between the biomedical image security, embedding capacity as well as image quality that can play significant role in telemedicine applications.

## VI. CONCLUSION

This paper has presented a novel and robust steganography technique for medical image steganography to be used for telemedicine application. Considering the inherent medical image characteristics and critical associated information, a discrete Ripplet transform (DRT) based message embedding model was developed, where patient's details were embedded in transform domain. Unlike, wavelet transform technique the ability of DRT to deal with singularity issues and image decomposition strengthened the proposed work to enable efficient LSB message embedding which eventually led higher imperceptibility and embedding capacity while maintaining better image quality. To further strengthen the imperceptibility and embedding capacity, adaptive genetic algorithm based optimal pixel adjustment process was performed. In addition, to alleviate the issue of statistical attack such as RS attack in channel, AGA based block flipping and adjustment model was developed, where the proposed system has exhibited better outcome and RS resiliency. This paper incorporated dual model security feature by applying RSA based secret data encryption and DRT and AGA enriched steganography that as a combined model has exhibited better PSNR, embedding capacity, imperceptibility and attack resiliency. These all novelties support it to be used for real time telemedicine purposes. In future, other type of Ripplet transforms such as RT Type II and RT Type-III can be explored to have better efficiency.